\documentclass[conference]{IEEEtran}
\IEEEoverridecommandlockouts
\usepackage{cite}
\usepackage{amsmath,amssymb,amsfonts}

\usepackage
{subfigure}
\usepackage{algorithm}
\usepackage{algpseudocode}
\usepackage[export]{adjustbox}
\usepackage{float} 
\usepackage{graphicx}
\usepackage{textcomp}
\usepackage{xcolor}

\usepackage{amsmath}  
\DeclareMathOperator*{\argmin}{arg\,min}
\def\BibTeX{{\rm B\kern-.05em{\sc i\kern-.025em b}\kern-.08em
    T\kern-.1667em\lower.7ex\hbox{E}\kern-.125emX}}
\begin{document}

\title{Irregularity Inspection using Neural Radiance Field\\
}

\author{

\IEEEauthorblockN{1\textsuperscript{nd} Tianqi(Kirk) Ding}
\IEEEauthorblockA{\textit{Electrical and Computer Engineering} \\
\textit{Baylor University}\\
Waco, Texas 76798 \\
kirk\_ding1@baylor.edu}\\

\IEEEauthorblockN{2\textsuperscript{nd} Dawei Xiang}
\IEEEauthorblockA{\textit{Computer Science and Engineering Department} \\
\textit{University of Connecticut}\\
Storrs, Connecticut 06269 \\
ieb24002@uconn.edu}
}

\maketitle

\begin{abstract}
With the increasing growth of industrialization, more and more industries are relying on machine automation for production. However, defect detection in large-scale production machinery is becoming increasingly important. Due to their large size and height, it is often challenging for professionals to conduct defect inspections on such large machinery. For example, the inspection of aging and misalignment of components on tall machinery like towers requires companies to assign dedicated personnel. Employees need to climb the towers and either visually inspect or take photos to detect safety hazards in these large machines. Direct visual inspection is limited by its low level of automation, lack of precision, and safety concerns associated with personnel climbing the towers. Therefore, in this paper, we propose a system based on neural network modeling (NeRF) of 3D twin models. By comparing two digital models, this system enables defect detection at the 3D interface of an object.
\end{abstract}


\section{Introduction}
Large outdoor machinery is exposed to environmental factors every day, leading to misalignment and damage to screws and wires on the machinery's surface. Currently, most companies rely on specialized personnel to inspect these machines. These personnel climb to high places, take photographs, and subjectively assess maintenance needs based on their personal experience. This method is not only inefficient but also highly subjective, bringing uncertainties and risks associated with employees climbing to high places. Therefore, there is an urgent need for a computer vision-based approach to efficiently and objectively detect defects.

The concept of the digital twin was initially introduced by Dr. M. Grieves [1], defining it as a three-dimensional model encompassing a physical product, a virtual product, and the interconnections between them. Digital twin technology is currently employed across various industries, including healthcare, energy, and manufacturing. In this paper, we propose a methodology for Irregularity Inspection based on the Digital Twin concept. Utilizing a UAV (Unmanned Aerial Vehicle), we captured images of the intact, standard drilling rig and generated a model using Neural Radiance Fields (NeRF). Additionally, we captured images of the drilling rig in the field using a drone to form a model with Neural Radiance Fields (NeRF), representing a model with potential defects. Ultimately, we conduct a comparative analysis of the two models to identify and detect any defects.

\section{Related Work}

Neural Radiation Field (NeRF) utilizes neural networks to learn the 3D structure of a scene from a 2D image, and then uses the learned knowledge to create realistic 3D spaces from different viewpoints in the original image. Compared to traditional 3D reconstruction techniques that require the storage of large point clouds or meshes, NeRF can efficiently utilize memory, which gives NeRF an advantage in storing complex large-scale scenes. Most importantly, in the face of complex lighting conditions (e.g., reflections), traditional 3D reconstruction requires a lot of manual intervention, but NeRF can cope with such problems and generate highly realistic 3D models[2,10,11]. 

The NeRF approach operates by utilizing a deep neural network to map 3D coordinates to a radiance value and a volume density. This network is trained with a collection of 2D images of a scene captured from different angles, with the corresponding 3D locations and viewing directions as inputs. Through this training process, the network learns to predict the volume density and color of the scene at different 3D points, essentially learning a continuous representation of the scene’s 3D structure and appearance.[3,8,9]

\begin{figure*}
\includegraphics[width=0.95\textwidth]{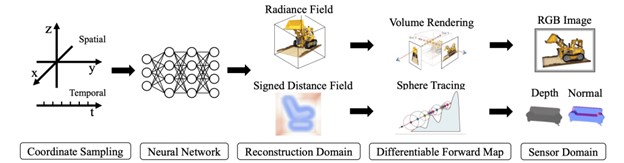}
\caption{Schematic representation of the NeRF procedure.[3]}
\label{fig}  
\end{figure*}

During the rendering phase, the 3D scene is reconstructed by querying this neural network at multiple 3D points along the light rays projected from the camera viewpoint through each pixel in the image. The color and bulk density predictions accumulated along each ray are then used to generate a new 2D image of the scene from the desired viewpoint, which produces a high-fidelity 3D reconstruction.[4]

\section{Methods}
\subsection{System Architecture}
We designed an intelligent system flow, as shown below, to realize an accurate and efficient AI Irregularity Inspection system.

\begin{figure*}
\includegraphics[width=0.95\textwidth]{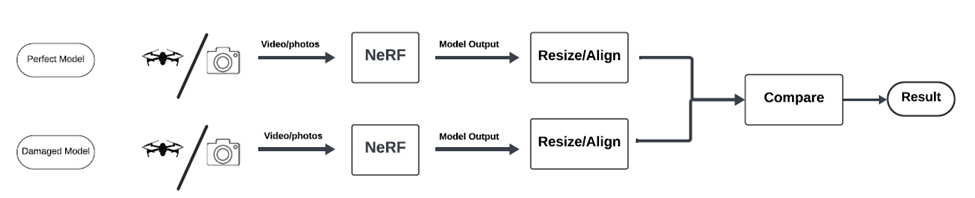}
\caption{AI Based Irregularity Inspection System Flowchart.}
\label{fig}  
\end{figure*}

As shown in the figure, the whole process starts with capturing images of the target using a drone or camera. Initially, we take two sets of images: the first set of photos captures a standard model of the target, which serves as a reference for future detection; the second set of photos records the current state of the facility. In subsequent steps, we process these two sets of images by feeding them into a Neural Radiation Field (NeRF) model to create two different 3D models, while the first set can be retained at all times. Immediately afterwards, we used the Iterative Closest Point (ICP) algorithm [5]to automatically align these models. This is because the size of the two models and their orientation will be different even though they are created from the same NeRF model. In the final step, the algorithm calculates the distance between the point clouds and sets a maximum threshold to mark mismatched regions to highlight any differences between them, ultimately generating a composite model that clearly identifies defects at overlapping points.

\subsection{Environment And Hardware Devices}\label{AA}
\paragraph{Software} We use NeRFstudio to enable Neural Radiation Field (NeRF) to generate 3D models. NeRFstudio is similar to the Neural Radiation Field (NeRF) toolset in that it provides access to NeRF and its extensions for modeling[6]. For model comparison, our system mainly uses open source libraries: Python 3.10, Open3D and NumPy. In the computer we use conda to manage these libraries and all the code runs on Visual Studio 2019.

\paragraph{Hardware} For NeRF model calculations we used an NVIDIA GeForce RTX 2060 Ti GPU to aid in parallel computing. For data acquisition we used a Canon EOS Rebel T4i camera(Fig. 3).

\begin{figure}
\includegraphics[width=0.3\textwidth]{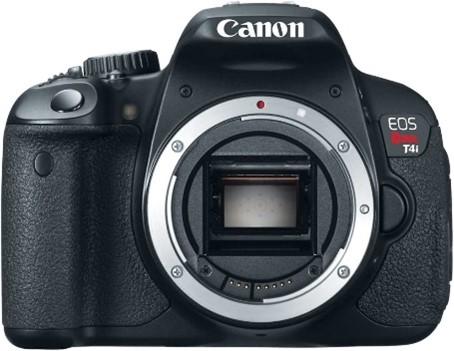}
\caption{Canon EOS Rebel T4i Camera.}
\label{fig}  
\end{figure}

\subsection{Iterative Closest Point and Point Cloud Comparison Algorithm}
\paragraph{Iterative Closest Point Algorithm} 
After acquiring the reconstructed models of both target and source objects via the NeRF process, a key problem is that the orientation and size of the two models are not the same.. Rather than imposing restrictions on input images to force uniformity, which could potentially undermine the generalizability of our method, we employ the Iterative Closest Point (ICP) algorithm to align the point clouds, ensuring they share the same orientation and size.

ICP[7]is a widely-used method for aligning two point clouds. And it has been pivotal in numerous 3D vision applications, demonstrating broad developmental prospects in areas such as space-based remote sensing, photogrammetry, and robotics. 

The ICP algorithm operates by iteratively minimizing the distance between the points of two point clouds. The algorithm begins with an initial guess of the transformation (rotation and translation) and iteratively refines the guess to find the best fit between the two point clouds. The algorithm consists of two main steps:
\begin{enumerate}
    \item	\textbf{Matching Step}: The algorithm identifies the closest points between the two point clouds. This is achieved using a nearest-neighbor search. In our implementation, we adopted the point-to-point distance as our chosen metric.
    \item \textbf{Minimization Step}: The algorithm computes the transformation that minimizes a cost function - in our implementation, this is the mean squared error between the points of the two point clouds.
\end{enumerate}


\begin{figure}[H]
\centering
\includegraphics[width=0.5\textwidth]{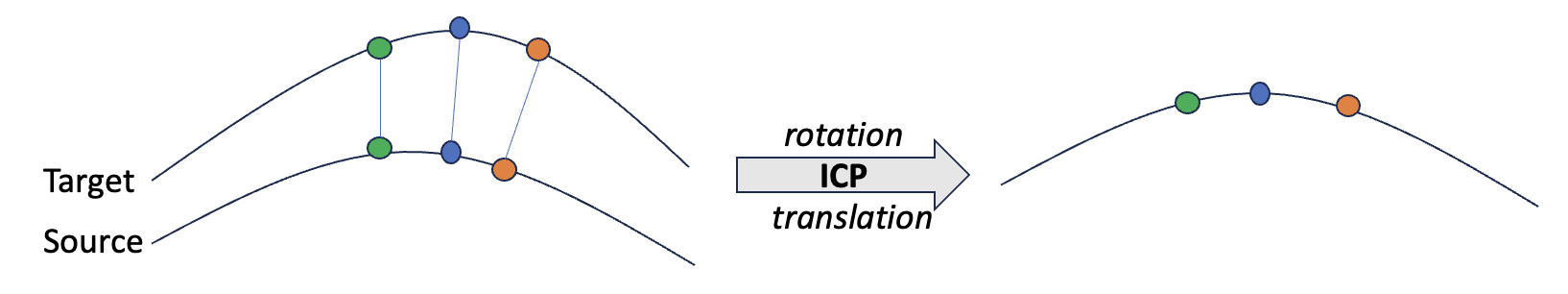}
\caption{A illustration of ICP algorithm. Our goal is to find a transformation to make source and target model as close as possible.}
\label{fig:icp}
\end{figure}


\begin{algorithm}
\caption{Iterative Closest Point (ICP)}
\begin{algorithmic}[1]
\Require Target point cloud $P$, Source point cloud $Q$, maximum iterations $maxIter$, and convergence tolerance $tol$
\Ensure Optimal transformation $T_{optimal}$
\State Initialize transformation $T$ to identity
\For{$i=1$ \textbf{to} $maxIter$}
    \hspace{\algorithmicindent}\Statex \hspace{\algorithmicindent}  First find closest points in $Q$ for each point in $P$ under transformation $T$, forming pairs $\{(p, q)\}$, then compute each point in $Q$ after transformation,  compute the total distance $E(T)$ for two point clouds, and we update $T$ to minimize the total distance.
    \State  $q^\prime = Tq$ 
    \State  $E(T) = \sum\limits_{(p,q^\prime)} |p - q^\prime| $, 
    \State $T^\prime = \argmin\limits_T (E(T))$, 
    \If{$||T^\prime - T|| < tol$}
        \State Break
    \EndIf
    \State $T = T^\prime$
\EndFor
\State $T_{optimal} = T$
\end{algorithmic}
\label{alg:icp}
\end{algorithm}

\paragraph{Point Cloud Comparison Algorithm} 

To compare the two aligned models, we use the distance between the closest point in the two models as a metric. For each point in the damaged model, the corresponding point in the perfect model is found via a nearest-neighbor search. Points are deemed 'matched' if the distance between the two points is less than a preset threshold. If this condition is not met, the point is deemed 'unmatched'.
Finally, we use the volume of the 'unmatched' portion as a measurement to determine if there is a substantial change in the original model. Fig. 9 depicts the 'unmatched' volume in pink.

\begin{figure}[h]
\includegraphics[width=0.45\textwidth]{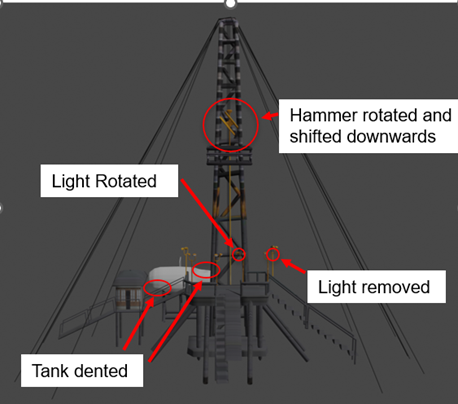}
\caption{Testing Tower Model.}
\label{fig:4a}
\end{figure}

\begin{figure}[h]
\includegraphics[width=0.45\textwidth]{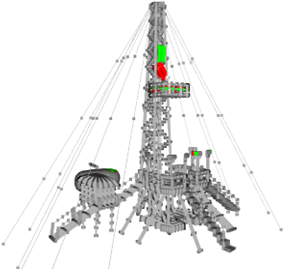}
\caption{Testing Tower Model after comparison }
\label{fig:4b}
\end{figure}

In order to verify the reliability of Comparison's code. We downloaded an electronic model of the virtual 3D Tower from the Blender website. We made some artificial changes to this model in three places by Blender software (see Fig \ref{fig:4a}). By running the Comparison code, we can see from Fig \ref{fig:4b} that the Comparison code marks all the changes correctly.

\section{Experiments}
Lego blocks are very common items on the market. We chose a very small Shiba Inu block toy for the first set of experiments (only the size of an adult's fist; see Fig 5a). This experiment is to test whether the 3D model generated by NeRF can effectively perform defect detection in the case of small objects. We took the first set of data with the Shiba Inu block retaining the tail (Fig 5a) and the second set of data with the Shiba Inu block removing the tail (Fig 5b), respectively, and generated two models through NeRFstudio (Fig 6 a\&b). The point cloud files of the two models were imported into the ICP code system. Since the two models were trained from the NeRF neural network twice, the size and orientation of the exported models were different. So our first step is to automatically adjust the model size and orientation by the code ICP system, then Align the two models (Fig 7), and then perform the Irregularity Inspection to get the final result (Fig 8).

\begin{figure*}[!t]
  \centering
  \subfigure[Shiba Inu block(with tail)  ]{
  \includegraphics[width=0.45\textwidth]{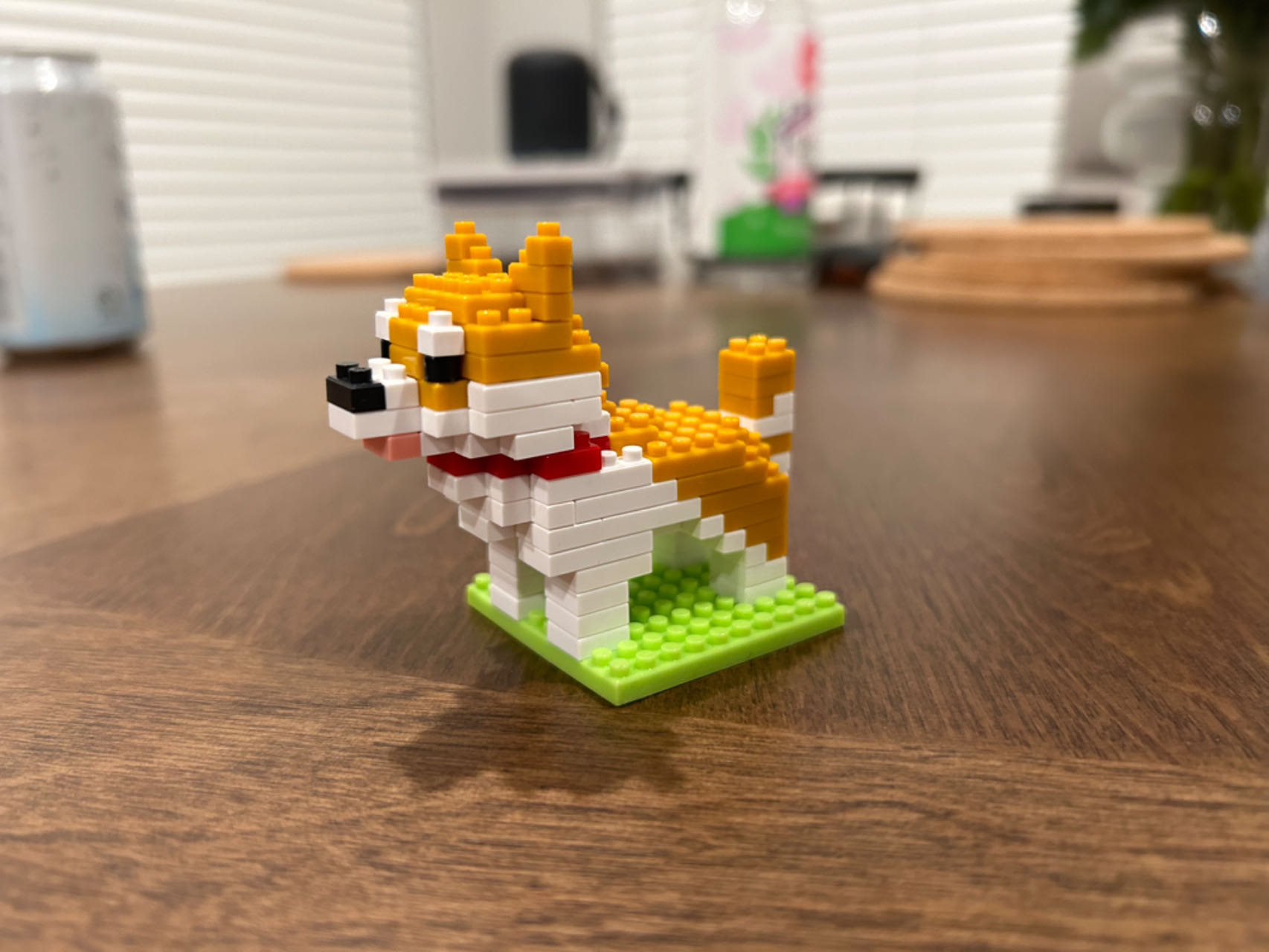}
    \label{fig:5a} }
  \subfigure[Shiba Inu block(without tail)]{
    \includegraphics[width=0.45\textwidth]{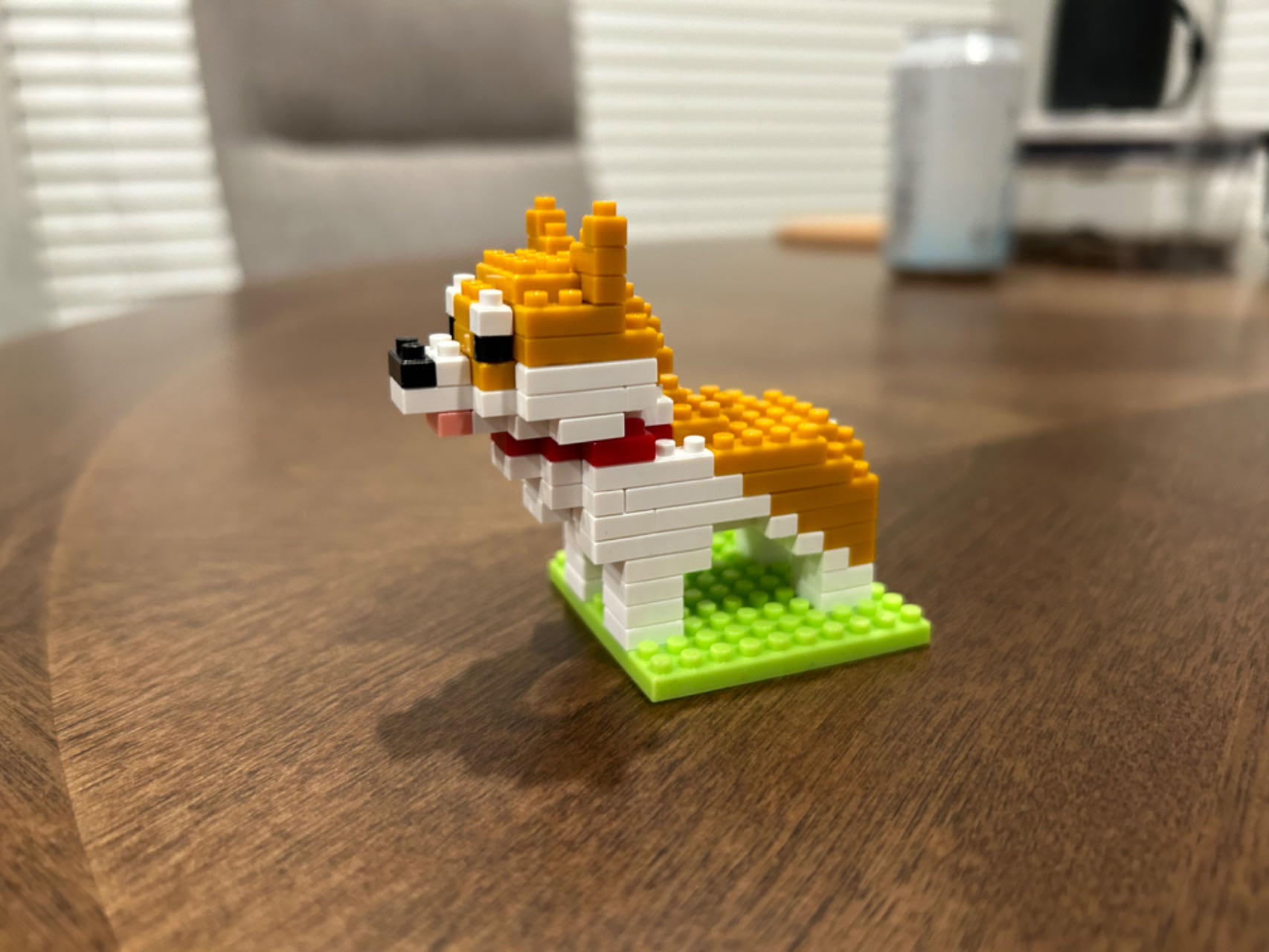}
    \label{fig:5b}
  }
  \caption{We use Shiba as the object}
  \label{fig:label_for_whole_figure}
\end{figure*}

\begin{figure*}[!t]
  \centering
  \subfigure[Shiba Inu NeRF model(with tail)]{
    \includegraphics[width=0.45\textwidth]{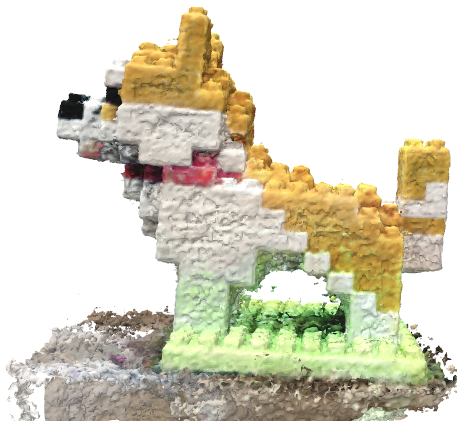}
    \label{fig:6a}
  }
  \subfigure[Shiba Inu NeRF model(without tail)]{
    \includegraphics[width=0.45\textwidth]{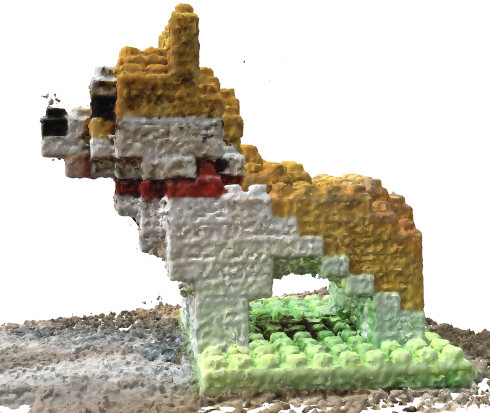}
    \label{fig:6b}
  }
  \caption{NeRF generated from source images}
  \label{fig:label_for_whole_figure}
\end{figure*}

\begin{figure*}[!t]
  \centering
  \subfigure[Shiba Inu NeRF model(with tail)]{
    \includegraphics[width=0.45\textwidth]{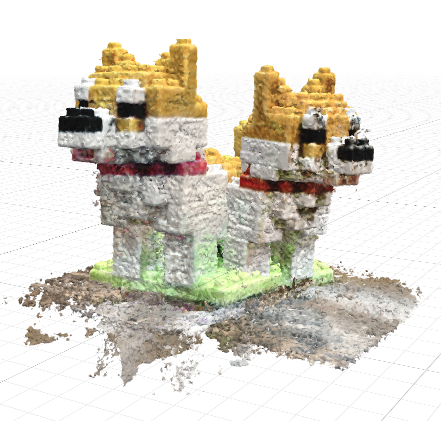}
    \label{fig:7}
  }
  \subfigure[Shiba Inu NeRF model(without tail)]{
    \includegraphics[width=0.45\textwidth]{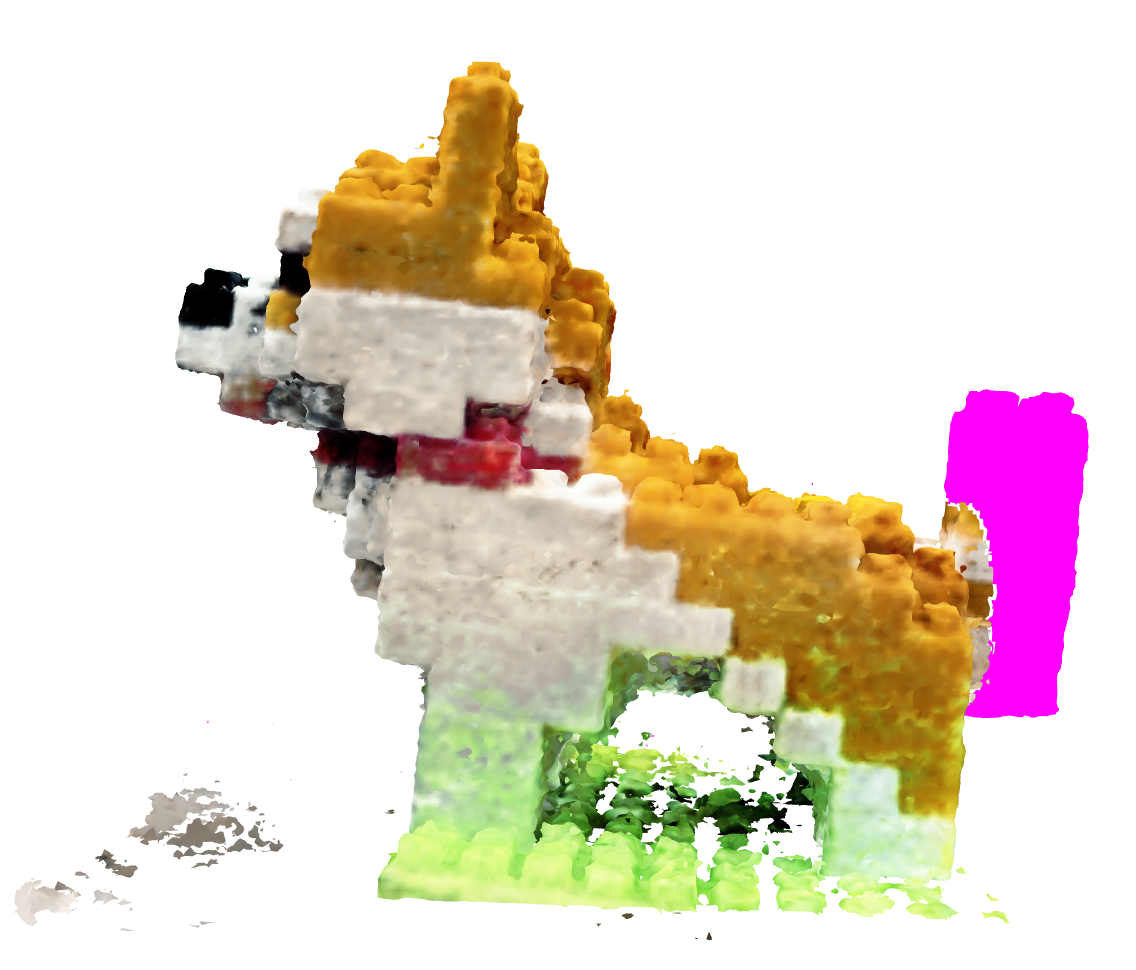}
    \label{fig:8}
  }
  \caption{Reconstructed result illustration}
  \label{fig:label_for_whole_figure}
\end{figure*}

\begin{figure*}[!t]
  \centering
  \subfigure[Before lifting the arm of the chair]{
    \includegraphics[width=0.45\textwidth]{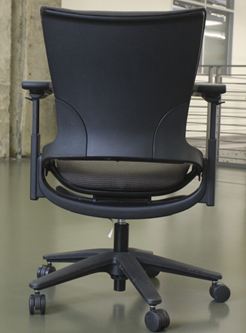}
    \label{fig:9a}
  }
  \subfigure[After lifting the arm of the chair]{
    \includegraphics[width=0.45\textwidth]
    {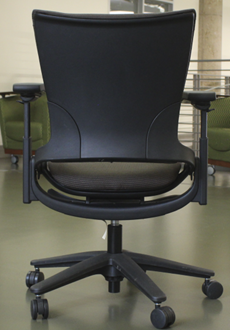}
    \label{fig:9b}
  }
  \caption{We also used chair as the second test cases}
  \label{fig:label_for_whole_figure}
\end{figure*}

\begin{figure}[h]
\includegraphics[width=0.4\textwidth]{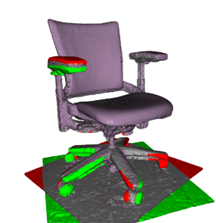}
\caption{Illustration of our model tested on chair. Green shows the parts appear before lifting the arm, red shows the part after lifting the arm. Our pipeline can clearly catch  the changed part.}
\label{fig:10}
\end{figure}

For the second set of experiments, we chose a laboratory chair to simulate the defects by varying the height of the chair armrest (Fig. 9). Here we add some colors to describe the differences between the two models after comparing them. red marks differences in the chair with the armrest up. green marks differences in the chair with armrest down. gray and purple areas mark where changes were smaller than the noise threshold. purple areas mark where changes were smaller than the noise threshold. The final results are very satisfactory.

\section{Conclusion}
The Neural Radiation Field (NeRF)-based Automated Integrated Irregularity Inspection System reduces the problem of traditional scanning models consuming large amounts of memory, optimizes the problem of excessive human interference required for models under complex lighting conditions, and the system generates models faster, giving those who perform Irregularity Inspection under different conditions more options. The system generates models faster, giving more choices to those who perform Irregularity Inspection under different conditions.

The main contribution of this system: For the first time, NeRF-based rendering technology is successfully used for the Irregularity Inspection task. This task provides more flexibility in maneuvering than traditional Irregularity Inspection, and requires less memory on the computer, providing a clearer 3D model.

Future work:
Apply the system to large-scale industrial scenarios, using drone sampling, and use that sample to test the accuracy of our system and neural rendering. Also in future work, we intend to seek more realistic models in an integrated manner, fusing traditional modeling methods with neural rendering methods to obtain more efficient and accurate models.


\end{document}